\documentclass{article}
\usepackage{Settings/ECRRS_packages}
\usepackage{Settings/ECRRS_Layout}
\usepackage{array}
\usepackage{graphicx}
\usepackage{caption}
\usepackage{wrapfig}

\final{} 

\begin{document}

\title{Royal Reveals: LiDAR Mapping of Kronborg Castle, Echoes of Hamlet's Halls}


\author{
    Leon Davies \orcidlink{0009-0004-9904-1909}\affil{Loughborough University} 
    \and
    {Simon Sølvsten \orcidlink{0000-0002-2610-9894}\affil[2]{European Center for Risk \& Resilience Studies}}
    }

\date{
    \today}

\makeatletter
\let\mytitle\@title
\let\myauthor\@author
\let\mydate\@date
\makeatother

\maketitle

\begin{mdframed}[backgroundcolor=gray!10]

    \begin{abstract}
    
    This paper presents a large scale dataset from a meticulous 360-degree LiDAR (Light Detection and Ranging) scan conducted on Kronborg Castle, a renowned Renaissance fortress located in Elsinore (Helsingør), Denmark, famously associated with Shakespeare's "Hamlet." Utilising a vertical mounted, gimbal stabilised, 16 channel, 360-degree Velodyne VLP-16 LiDAR scanner, paired with an Intel RealSense L515 depth camera. This research offers an unparalleled digital representation of the castle's intricate architectural details and structural nuances, enabling fellow researchers to conduct experiments utilising the data for SLAM (Simultaneous Localisation and Mapping) as well as floorplan generation.
    
    \end{abstract}

\end{mdframed}


\begin{multicols}{2}
   
    \section{Introduction}   \label{Introduction}
        
        Publicly available large-scale real-world indoor point cloud surveys of LiDAR data are a rare resource. This rarity can cause difficulty within research projects that utilise this data. This is likely due to the high cost of data, not just in the monetary sense of the sensors themselves but also in the time required to plan, navigate and map a building. Obtaining access and permission to run sensors within large indoor buildings can also cause privacy concerns. Due to these reasons, there is a scarcity of data on large indoor LiDAR scans. Previous datasets, such as the Radish Dataset \cite{Radish}, are no longer supported and have become unusable in their original format with modern-day software and SLAM systems. Other indoor datasets such as \cite{ScanNet,song2017,choi2015robust} focus on smaller environments of individual rooms or scenes, which limit their usability for large-scale SLAM experiments. Many datasets of individually scanned objects exist in PointCloud format \cite{qi2016pointnet}, which enable tasks such as PointCloud-based object recognition; this differs from our dataset as our insight is that we can conduct more accurate object recognition through RGB input. We align our PointClouds with RGB video stream to enable object recognition to be used for downstream tasks such as Semantic SLAM or automated floor plan creation. We provide pointclouds built through SLAM using \cite{dlo} of each individual floor of the castle as well as an entire pointcloud for the entire site. We also provide 2D representations of the castle built through occupancy grid mapping. 
        
    
    \subsection{About Kronborg Castle}   \label{Kronborg Castle}
    
        Kronborg Castle is situated in the Danish city of Elsinore (Helsingør) on the island of Zealand, located to the north of Copenhagen. While the castle is likely most famously associated with Shakespeare's "Hamlet," it stands as a prominent Renaissance castle, earning its place on the UNESCO World Heritage list in 2000 (\cite{403324385, UNESCOKronborg}).
    
        Established around 1420 by Erik of Pomerania, also recognized as King Erik VII, Kronborg Castle's precursor, known as Krogen, strategically stood at the narrowest passage between Denmark and Sweden. This strategic location enabled Krogen, and subsequently Kronborg, to levy tolls on Sound's (Øresund's) maritime traffic, strengthening the Danish Crown's control over the Baltic Sea trade routes and solidifying Denmark's influence in Northern Europe (\cite{403324385}).
    
        Over the centuries, the castle has undergone numerous modifications and reconstructions. Particularly noteworthy is the period between 1574 and 1585 when King Frederik II spearheaded a substantial overhaul of the original Krogen, shaping it into the iconic Kronborg we recognize today. This ambitious project aimed not only to fortify the castle's structure and bolster its defensive prowess but also to establish it unequivocally as a symbol of power (\cite{Grinder-Hans2018, 403324385}).

        The Kronborg Castle comprises a basement historically used for storage, kitchens and other utility purposes, a ground floor covering which contained public rooms, such as the Great Hall, reception rooms and the Chapel, which was for private apartments and bedrooms for the nobility, a second floor measuring with additional private rooms for guest accommodation, and a third floor, which encompasses the attic area, which was likely used for storage or servants' quarters.

        Our dataset encompasses a comprehensive scan of the building, with the exception of 3 rooms to which we could not gain access. 

        For readers interested in visual representations, a photograph of the castle is available in Figure \ref{fig:castle_pic}.

                    \begin{minipage}{\linewidth}
    \centering
    \includegraphics[width=0.37\linewidth]{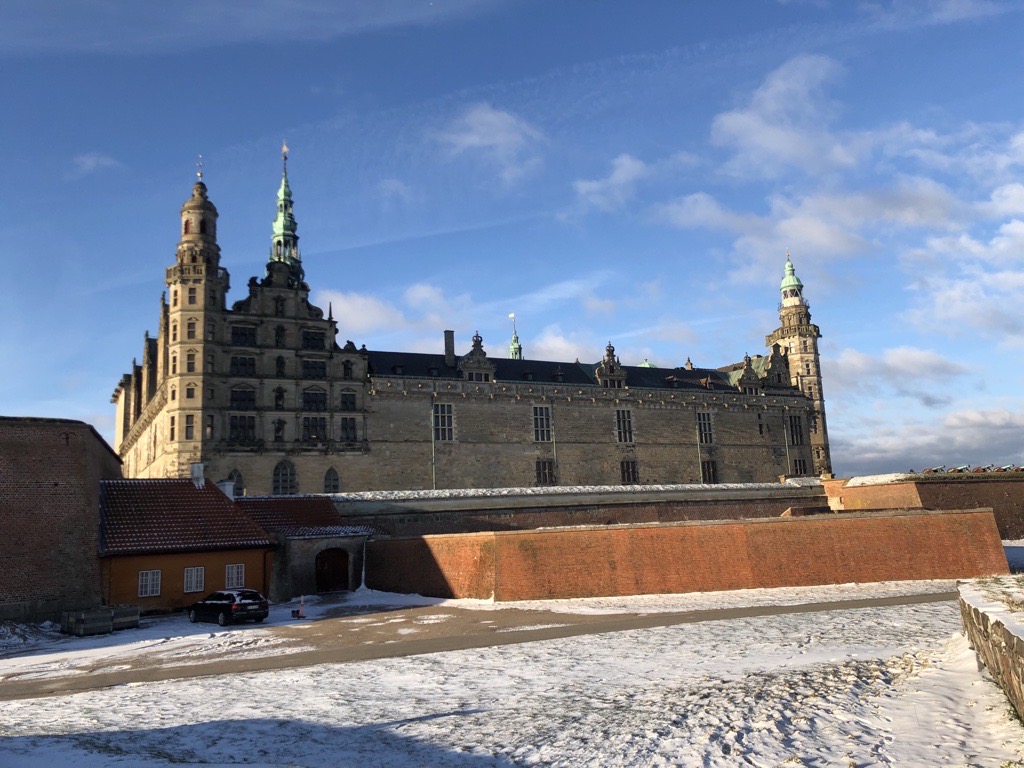}
    \captionof{figure}{A photograph of Kronborg Castle.}
    \label{fig:castle_pic}
\end{minipage} 
        
        Please refer to Table \ref{tab:About_Kronborg_Castle} for a summary of the provided information about Kronborg Castle.
    
        \end{multicols}
            \begin{table}[h]
    \centering
    \caption{About Kronborg Castle}
    \label{tab:About_Kronborg_Castle}
    
    \begin{tabularx}{\textwidth}{lX}
        \toprule
        \textbf{Year} & \textbf{Historical Events} \\
        \midrule
        1400-1420 & Krogen Castle was built, replacing an earlier royal castle located further inland. \\
        1574-1585 & Reconstruction and modernisation of Kronborg Castle under King Frederik II. \\
        1629-1639 & Renovation of the castle after a significant fire occurred under King Christian IV. \\
        1658-1660 & The castle suffers significant destruction during the war with the Swedes. \\
        1785 & Kronborg Castle undergoes modifications to serve as military barracks. \\
        1915 & Establishment of the Trade and Maritime Museum within Kronborg. \\

        \toprule
        \textbf{Type} & \textbf{Building Characteristics} \\
        \midrule
        Address & Kronborg 1B,  \\
        City & Helsingør (Elsinore) \\
        Zip code & 3000 \\
        Latitude, Longitude & 56.04016758452486, 12.622483167867747 \\
         Dimensions & Length: 145M, Width: 74M \\
         Tallest Point & 62M (Telegraph Tower)\\
         Total Floor Area & 16,000$^M$ \\
        \bottomrule
    \end{tabularx}
\end{table}
        \begin{multicols}{2} 
        
    \section{Methodology}        \label{Methodology}
    
        The data collection process utilsed a 360-degree vertically mounted and gimbal-stabilised VLP-16 Velodyne LiDAR scanner, complemented by an Intel RealSense L515 LiDAR depth camera affixed directly below the LiDAR scanner. Both sensors were securely mounted on a DJI RS 3 Pro Gimbal to ensure optimal stability and versatility during the scanning process.

        Further enhancing our setup for precision, the gimbal was subsequently attached to a tripod equipped with wheels. This arrangement was designed so the gimbal minimised vertical movement, while the tripod served the dual purpose of reducing potential shaking from walking motions and maintaining a consistent height for the sensor mount.

        Despite meticulous planning and execution to attain the highest data quality, it's essential to acknowledge the inherent challenges encountered during the scanning process. Navigating around Kronborg Castle presented various obstacles that occasionally affected the sensor's height and angle. Moreover, certain conditions intermittently induced vibrations in the sensors. While these challenges introduced some variability, our comprehensive efforts aimed to mitigate these factors and yield a dataset of commendable quality for scholarly analysis.
    
        The configuration and mounting setup, as described, are further shown in figure \ref{fig:gimbal} for enhanced clarity and visualisation.

        Data was recorded in bag format through ROS to enable future work to utilise the data temporally, we provide ROS bags for the entirety of Kronborg Castle split up into 13 seperate files. We also provide ROS2 bags, as converted by rosbags.

        The decision to partition the mapping sessions was done due to the sheer size of the castle and the difficulty of maneuvering around between different sections. 

        A breakdown of each recording is available in Table \ref{tab:Scan Breakdown}

            \begin{minipage}{\linewidth}
    \centering
    \includegraphics[width=0.37\linewidth]{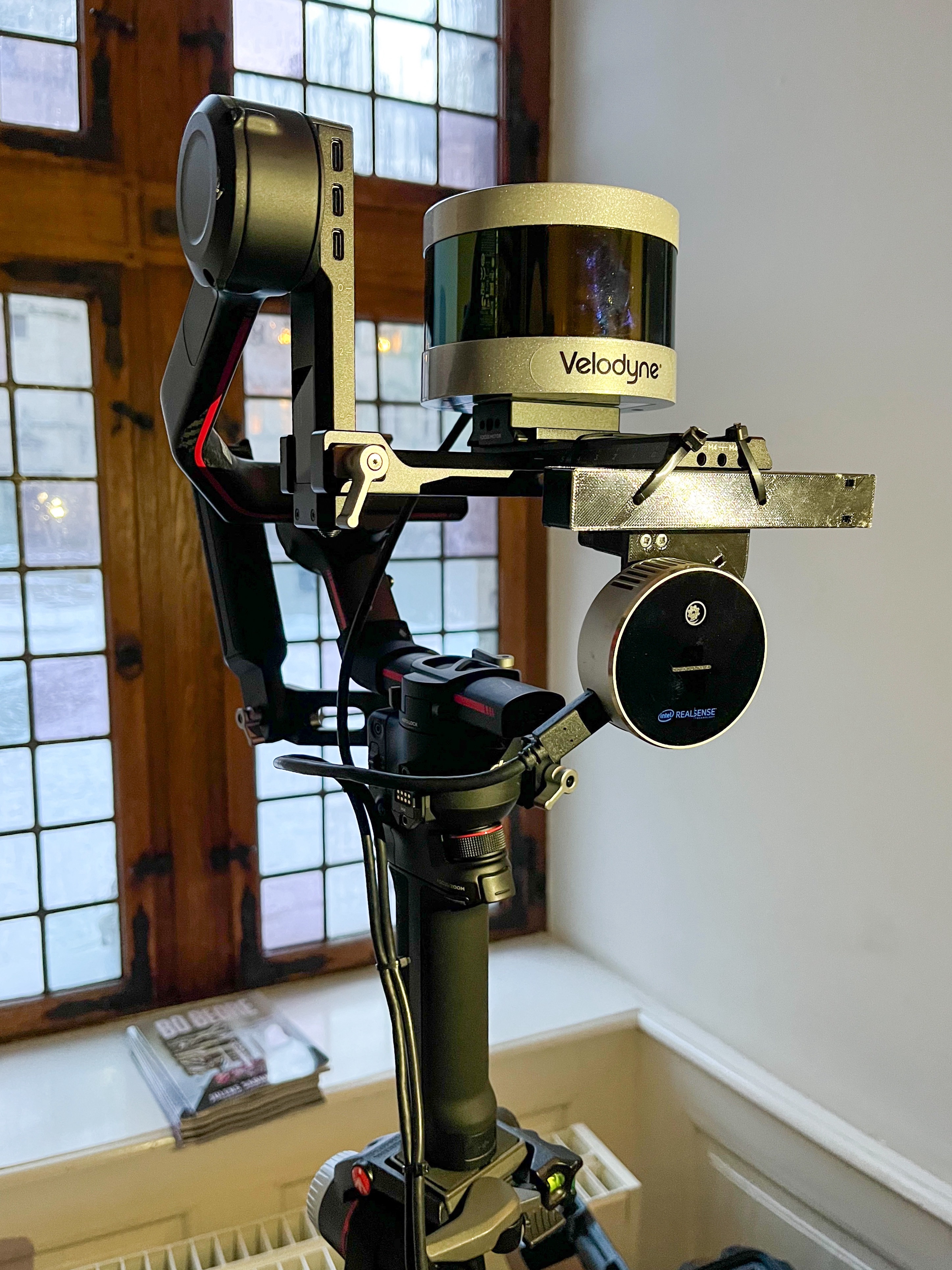}
    \captionof{figure}{Scanning apartus used to scan Kronborg Castle.}
    \label{fig:gimbal}
\end{minipage} 
        \end{multicols}
            \begin{table}[h]
    \centering
    \caption{Scan Breakdown}
    \label{tab:Scan Breakdown}
    
    \begin{tabularx}{\textwidth}{lX}
        \toprule
        \textbf{File} & \textbf{Description} \\
        \midrule
        Kronoborg-1 & Ground floor, south eastern section and courtyard. \\
        
        Kronoborg-2 & Ground floor - north eastern section. \\
        
        Kronoborg-3 & Ground floor - west section. \\
        
        Kronoborg-4 & First floor - entire floor. \\
        
        Kronoborg-5 & Second Floor - entire floor.) \\
        
        Kronoborg-6 & Third floor - west, north and southern sections. \\

        Kronoborg-7 & Third floor - east section. \\

        Kronoborg-8 & Basement - west section. \\

        Kronoborg-9 & Basement - south section. \\

        Kronoborg-10 & Basement - north section. \\

        Tower-bottom & Turret tower bottom floor \\

        Tower-mid1 & Turret tower middle first floor \\

        Tower-mid2 & Turret tower middle second floor \\

        Tower-top & Turret tower top floor \\
    \end{tabularx}
\end{table}
        \begin{multicols}{2} 
    
    \section{Data}               \label{Data}

    We provide raw data in the form of bag files which contain temporal scans and RGB images. We also provide odometry data adjoining the pointclouds computed using \cite{dlo}. Additionally, we have maps computed by compiling the data in the forms of point clouds and occupancy grid maps. 

            \begin{minipage}{\linewidth}
    \centering
    \includegraphics[width=0.37\linewidth]{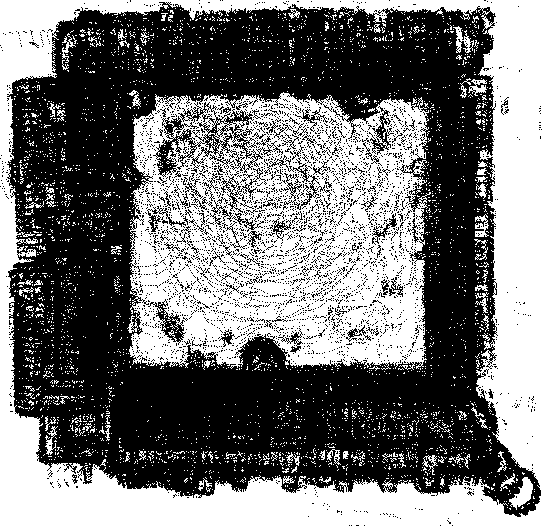}
    \captionof{figure}{Top down view of the entire point cloud of Kronborg Castle}
    \label{fig:entire_pcl}
\end{minipage}

As the mapping of the building was split up into several separate scans, we provide data of each separate recording as well as compiled data of the entire castle and separate floors.

            \begin{minipage}{\linewidth}
    \centering
    \includegraphics[width=0.55\linewidth]{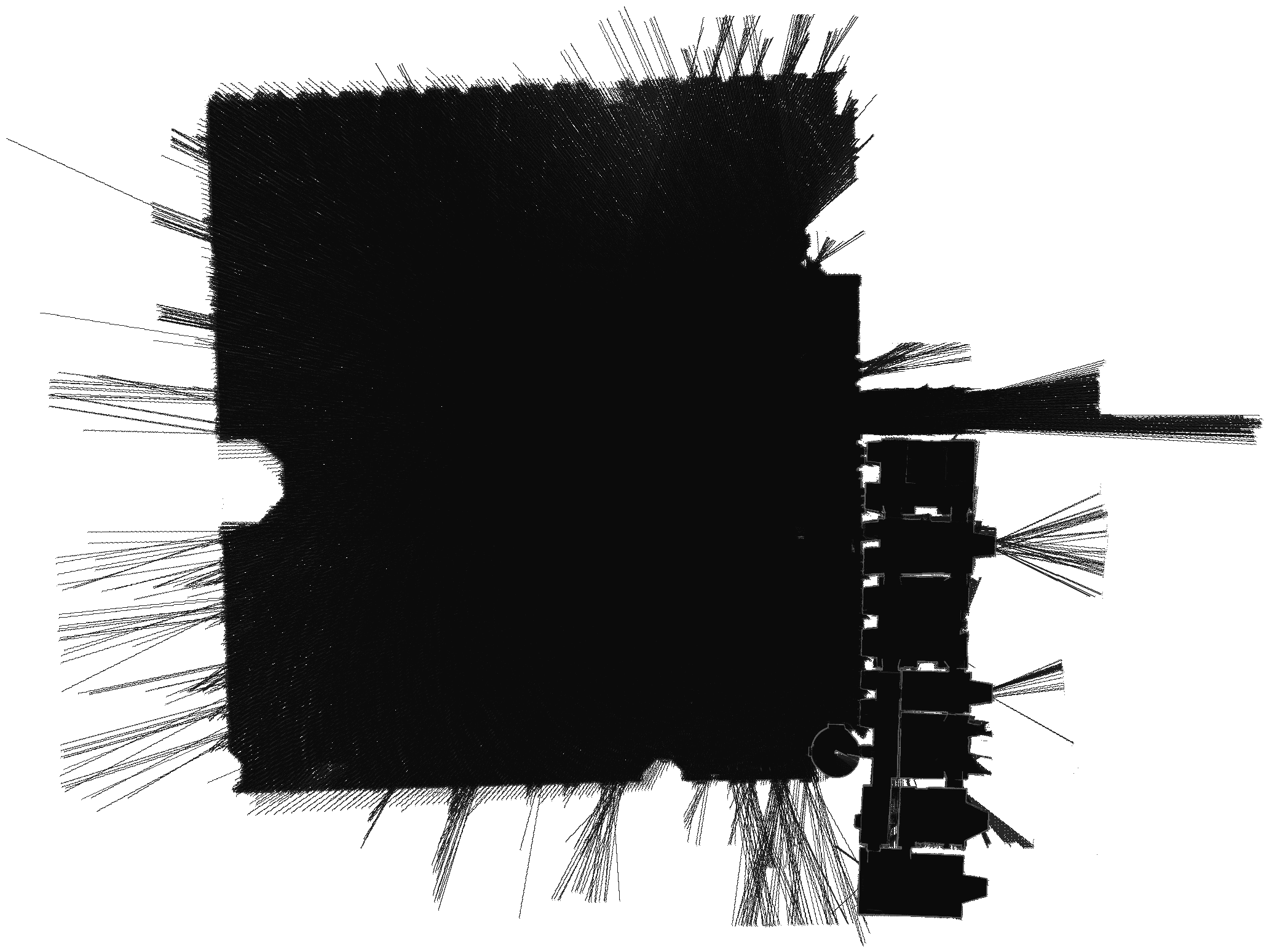}
    \captionof{figure}{Sample of a Grid map produced of the a section of the ground floor of Kronborg Castle.}
    \label{fig:ogm}
\end{minipage}

        \end{multicols}
            \begin{table}[H]
    \centering
    \caption{Data Specifications}
    \label{tab:Data_Specifications}
    
    \begin{tabularx}{\textwidth}{lX}
        \toprule
            \textbf{Formats} & 3D PointCloud (VLP-16), 3D Dense PointCloud (L515), 2D LaserScan, Depth Image, RGB Image \\
            \textbf{Sensors} & Velodyne VLP-16, Intel Realsense L515  \\
            \textbf{Accessories}  & DJI RS 3 Pro Gimbal, 3D printed custom mount for Intel RealSense L515 depth camera, Manfrotto 055 tripod and Manfrotto Basic Dolly 127  \\
            \textbf{Spatial or Temporal Elements} & The construction scans were conducted on January 4, 2024, the total recorded scan time was 6 hours across 13 separate recordings\\

        \bottomrule
    \end{tabularx}
\end{table}
            \begin{table}[H]
    \centering
    \begin{tabular}{|c|c|c|c|c|c|}
        \hline
        \textbf{Floor}  & \textbf{Basement} & \textbf{Ground Floor} & \textbf{First Floor} & \textbf{Second Floor} & \textbf{Third Floor (Attic)} \\
        \hline
        \textbf{PointCloud} & 
        \includegraphics[width=0.10\textwidth]{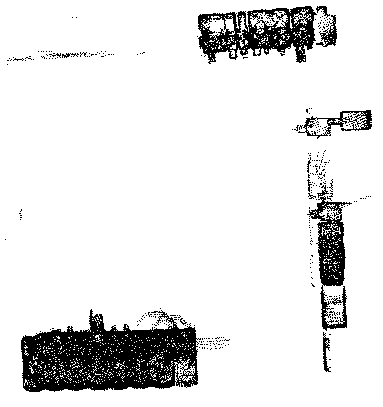} & 
        \includegraphics[width=0.10\textwidth]{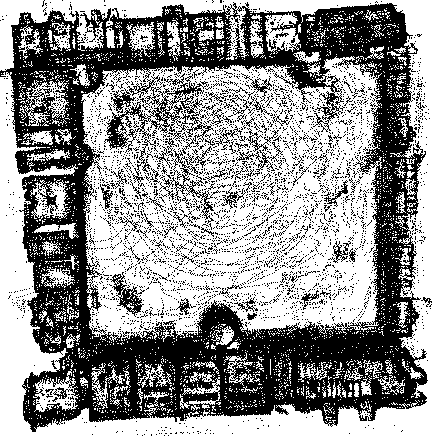} & 
        \includegraphics[width=0.10\textwidth]{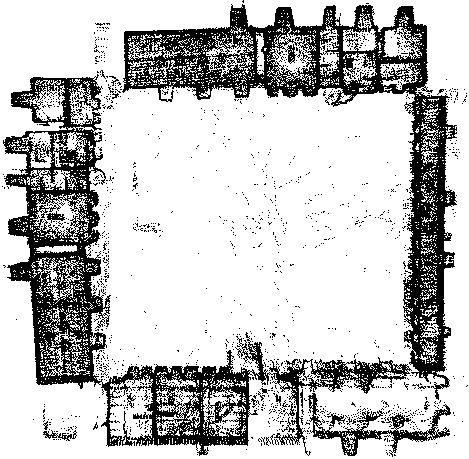} & 
        \includegraphics[width=0.10\textwidth]{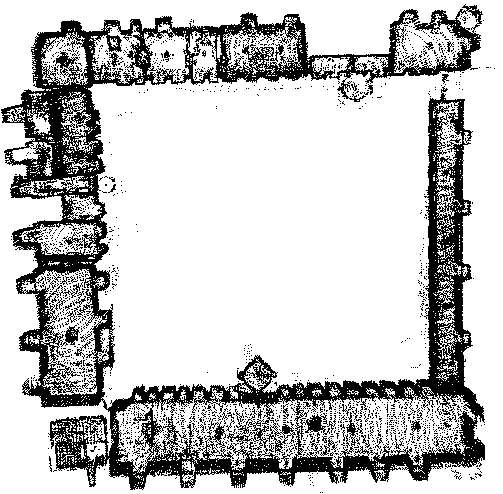} & 
        \includegraphics[width=0.10\textwidth]{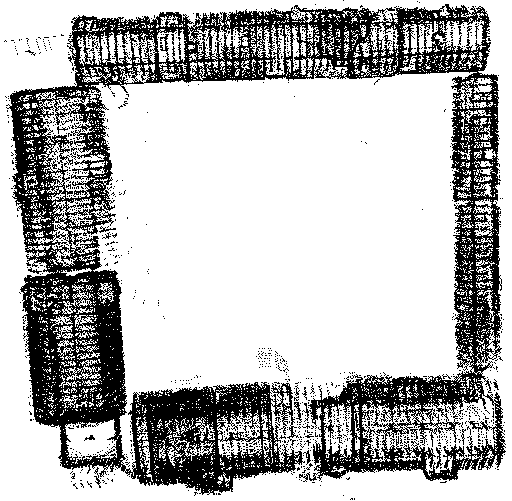} \\
        \hline
        \textbf{Floor Plan} & 
        \includegraphics[width=0.10\textwidth]{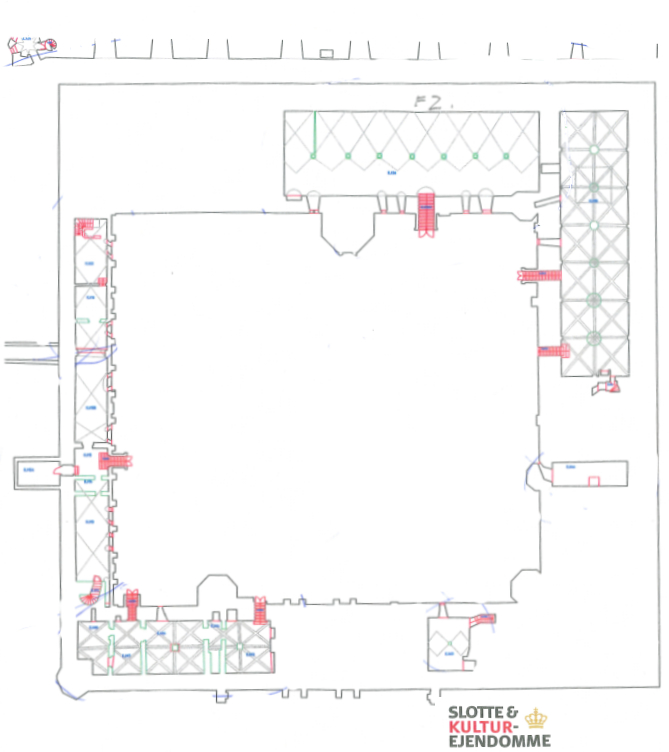} & 
        \includegraphics[width=0.10\textwidth]{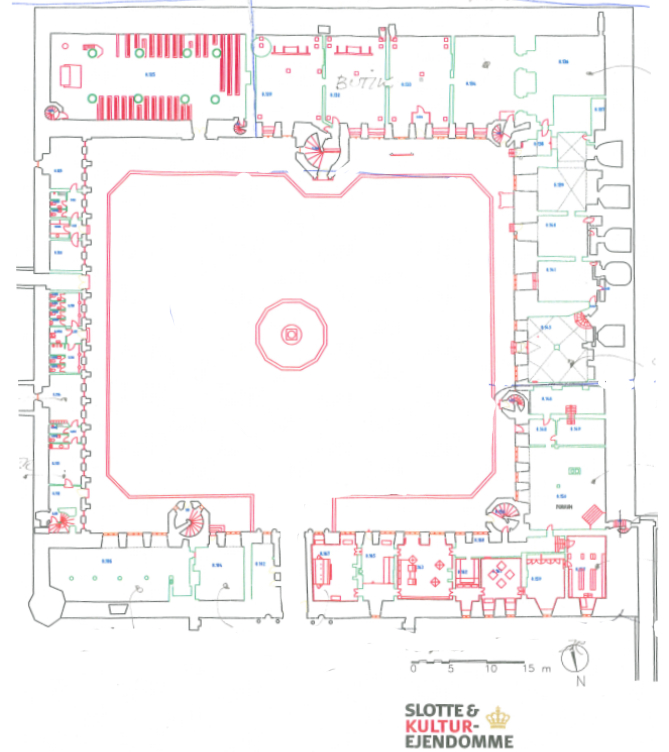} & 
        \includegraphics[width=0.10\textwidth]{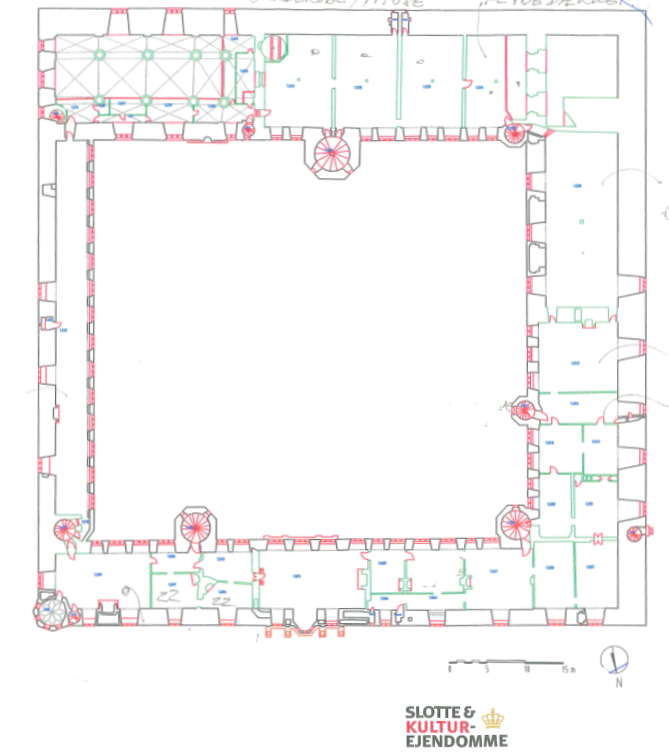} & 
        \includegraphics[width=0.10\textwidth]{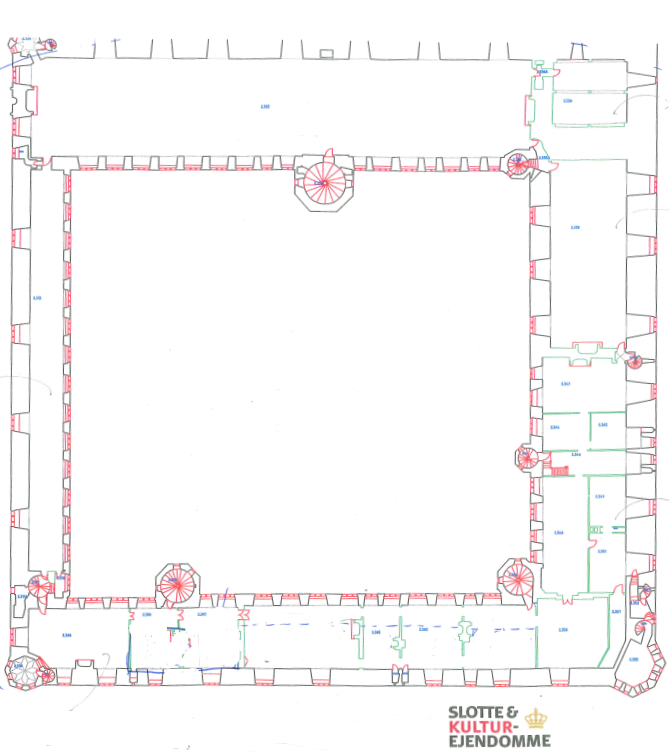} & 
        \includegraphics[width=0.10\textwidth]{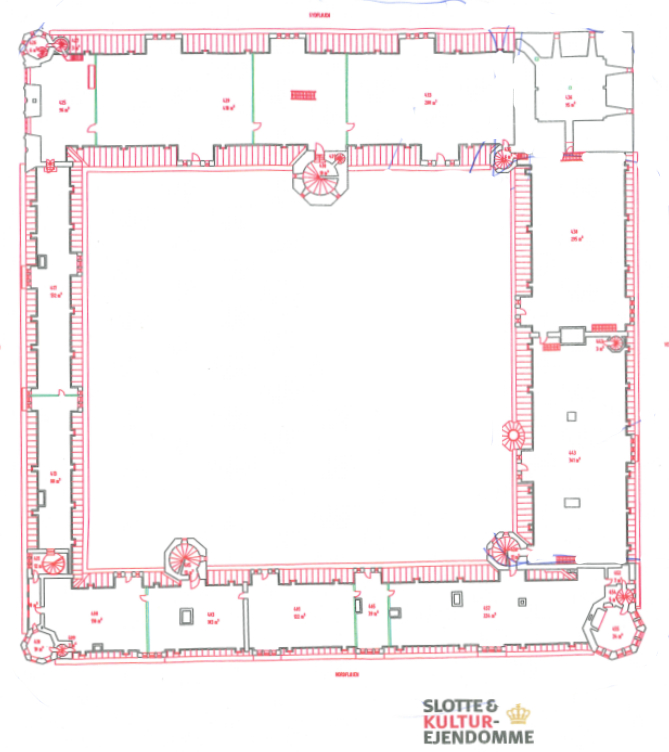} \\
        \hline
    \end{tabular}
    \caption{Kronborg Castle: Point Clouds}
    \label{tab:pointclouds}
\end{table}
        \begin{multicols}{2} 

        Access to download the dataset is available at http://www.github.com/bigggs/kronborg.
    
    \section{Acknowledgment}       \label{Acknowledgment}
    
        The authors express gratitude to the Danish Agency for Culture and Palaces for facilitating access to Kronborg Castle, pivotal to our study's success. Special thanks to the Kronborg Castle staff for their invaluable support, especially during challenges with the historical and, at times, stubborn locks.
        
        Additionally, we deeply appreciate Thomas Rahbek and Trine Neble from the Danish Agency for their unwavering support and trust. Their collective efforts significantly advanced our research objectives.
          
    \section{Funding}       \label{Funding}           
    
        The WTW Research Network, an official division of Willis Towers Watson (WTW), generously supported this research.
        
        The authors express profound appreciation to the WTW Research Network for its financial assistance, which was instrumental in facilitating the comprehensive data collection outlined in this paper. Importantly, this financial support did not lead to any conflicts of interest. Furthermore, the funding entity made no efforts to sway the research design, approach, outcomes, or interpretation of results.
    
        It is imperative to underscore that this study was conducted with complete autonomy, adhering rigorously to established academic and ethical standards. The insights and conclusions presented herein stem from an impartial analysis, ensuring unwavering scientific integrity and independence throughout the research process.

\end{multicols}
\bibliographystyle{apalike}
\bibliography{bibliography.bib}
   



\end{document}